\documentclass[runningheads]{llncs}
\usepackage[T1]{fontenc}
\usepackage{graphicx}
\usepackage[export]{adjustbox}
\begin{document}
\title{Music Recommendation System based on Emotion, Age and Ethnicity}
%
%
\author{Ramiz Mammadli\inst{1},
Huma Bilgin\inst{2}, \and Ali Can Karaca\inst{3}}

%
%
\institute{Department of Computer Engineering, Yildiz Technical University, Davutpa\c{s}a Campus, 34220 Istanbul Turkey\\
\email{\inst{1},\inst{2}{ramiz.mammadli,huma.bilgin}@std.yildiz.edu.tr, \inst{3}ackaraca@yildiz.edu.tr}\\
\url{https://ce.yildiz.edu.tr}}
\maketitle              
\begin{abstract}
A Music Recommendation System based on Emotion, Age and Ethnicity is developed in this study, using FER-2013 and ``Age, Gender and Ethnicity (Face Data) CSV'' datasets. The CNN architecture, which is extensively used for this kind of purposes has been applied to the training of the models. After adding several appropriate layers to the training end of the project, in total, 3 separate models are trained in the Deep Learning side of the project: Emotion, Ethnicity and Age. After the training step of these models, they are used as classifiers in the web application side. The snapshot of the user taken through the interface is sent to the models to predict the their mood, age and ethnic origin. According to these classifiers, various kind of playlists pulled from Spotify API are proposed to the user in order to establish a functional and user-friendly atmosphere for the music selection. Afterwards, the user can choose the playlist they want and listen to it by following the given link.
\keywords{Deep Learning \and CNN \and Emotion \and Age \and Ethnicity \and Web application \and Music Recommendation System \and Playlist.}
\end{abstract}
\section{Introduction}
People all over the world face a lot of stress in their daily life. This brings along the need of a solution to escape from it. The most effective and indispensable of these is music, which everyone in the world can easily access. It is no coincidence that more than 90\% of current world population listens to music an average of 32.1 hours per week \cite{article1}. This demand for music has naturally brought many entrepreneurial ideas with itself, and it continues to do so. 

As technology develops, users’ expectations also increase. For instance, expectations from a music streaming platform are not limited to just listening to musics. User-friendly interface, various playlists, music recommendation techniques etc. are now expected from applications. Therefore, in order to provide a better experience to users, personalized playlist generation and artist recommendation systems based on previously listened songs have been researched and used in many applications \cite{article2}. But as it can be foreseen, the user needs to use the application for a while and, sort of, introduce himself in order to benefit from these features. Moreover, the user may want to listen to various music genres at different states of emotion. This makes it complicated to guess from the training that the application has acquired based on the past. 

Many approaches have been developed for the recognition of human emotion \cite{EEG}, \cite{Wearable}. For example, features obtained from electroencephalograph (EEG) signals were used to emotion recognition during music listening \cite{EEG}. Additionally, in \cite{Wearable}, emotion of the user who uses music recommendation system was classified by a wearable device including galvanic skin response and photo plethysmography physiological sensors. Although these works provide an efficient way of discovering emotion, they are not practical for anyone that does not have these devices. More practical way to find human emotion class is that using speech and video data captured by microphones and web cameras \cite{AudioVideo}, \cite{SpeechVisual}. Here, the most part of music recommendation researches use only facial emotion expression from webcam images \cite{JoIEEE}, \cite{LSTM}, \cite{Opencv}, \cite{Mood}. Generally, these works use face detection and CNN-based classification models. In addition, several papers have been concentrated on developing the smart music players that use emotional recognition for mobile phones \cite{Smart}, \cite{Android1}. However, it is not a good idea to recommend the same playlist for a child and an adult person having the same emotion. Additionally, the recommendations may be different for the people who have different nationalities.

Considering such problems, we aimed to develop a music recommendation system by considering the emotion, age and ethnicity together from the instant picture taken from the user. To the best of our knowledge, it is the first work that uses these three different features for recommending music playlists. All these processes will be made accessible and usable by the end user through the web application. Therefore, it is aimed to provide the user with a better quality music experience.

\section{Data Sets}

In order to train the models that classify emotion, age and ethnicity of the user, we have used two datasets in this paper. The first dataset, which is used to detect the emotion of the user, is Facial Emotion Recognition 2013 Dataset (FER-2013) \cite{article3}. FER-2013 consists of 32,289 gray face images with 48x48 pixel dimensions. Feeding such a low resolutional gray image provides lower number of parameters in training models. Therefore, it is suitable for real-time applications about emotion recognition. There are 7 emotions in the dataset, however only 4 of them are used in the system which are angry, happy, neutral and sad. The remaining 3 emotions, which are disgust, fear and surprise are not used, since it is not so much realistic that the user is going to have that state of emotion when the instant picture is taken by the system. The data retrieved from the source given above is in .png format. Additionally, the dataset is already divided into two parts: train and test. It eliminates the step of split of the data to train and test subsets. 

The second dataset titled by ``Age, Gender and Ethnicity (Face Data) CSV'' is used to determine the ethnicity and age of the user \cite{ageData}. This dataset (AGE) is labeled by various tags. For instance, the tag, which is used for ethnicity is labeled from 0 to 4, representing White, Black, Asian, Indian, and Others (like Hispanic, Latino, Middle Eastern), respectively. Also, there are labels from 0 to 116, that indicate the age of the person in the image. There is also classification for gender, but that labelization is not going to be used in the system. The images are kept as pixels in 1 dimensional arrays. Therefore, to apply the models on them, they are pre-processed and transferred into 3 dimensional arrays. Note that we assume that users from the same ethnicity listen similar musical categories in this paper.

\section{System Analysis}
    The project consists of two main parts. In the first part, three CNN-based deep learning methods are implemented on the image obtained from the user's camera via the Web Application. By that way, the user's emotion, age and ethnicity status is determined in parallel. In the second stage, the appropriate playlist selection is made according to the features mentioned above. 
    
\begin{figure}[!h]
        \centering
        \includegraphics[scale = 0.44]{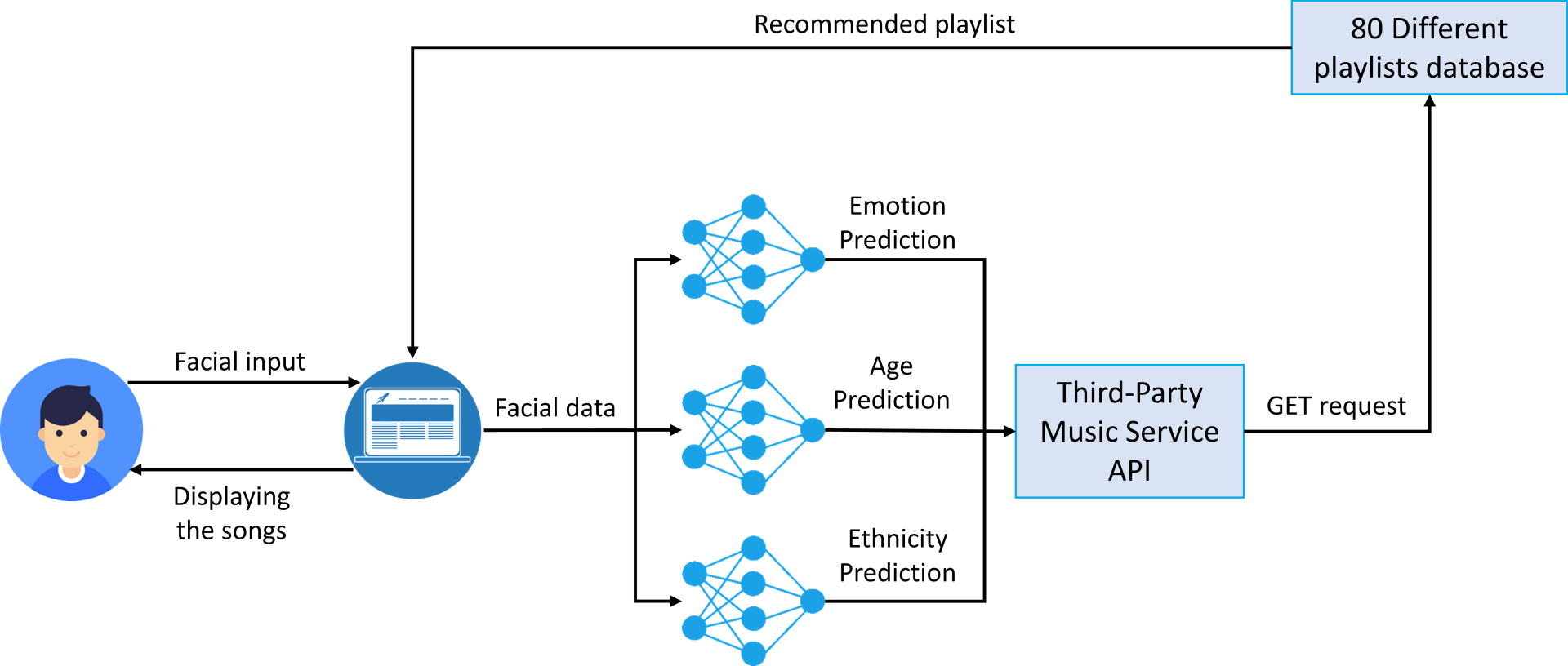}
        \caption{Block Diagram of the Proposed System}
        \label{fig:Use Case Diagram}
    \end{figure}

   General process of the system is shown in the Fig. \ref{fig:Use Case Diagram}. First, the user will access the application's web page. Then the user will be able to take an instant snapshot by activating their camera within the application. The image obtained from the camera will be processed by three parallel CNN models. Emotion, age and ethnicity will be determined on the frame taken, and a playlist suggestion will be made accordingly. Since there are 3 models, which determine emotion, age, and ethnicity separately, the predictions will be done simultaneously. All details about data processing phase are shared in the following section. 
          
\section{Data Processing phase}

    \subsection{Preprocessing}
    \subsubsection{Data Cleaning}
\begin{figure}[!h]
        \centering
        \includegraphics[scale = 0.405]{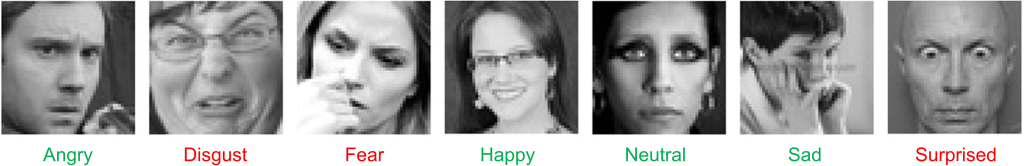}
        \caption{Samples from FER-2013 dataset \cite{article3}}
        \label{fig:sample_fer2013}
    \end{figure}

    Data cleaning is the process of detecting and correcting or deleting irrelevant, outlier, corrupt or incorrect data in a dataset. There are two main goals in this project. The first one is emotion detection which is developed using the FER-2013 dataset. The data set consists of seven different emotion classes. The ones required for our project are happy, angry, neutral and sad. For this reason, we discarded the photos of the remaining three classes, including fear, surprise, and disgust, from the dataset (Fig. \ref{fig:sample_fer2013}). Apart from that, since the used data is gathered professionally, excluding pre-mentioned outliers, no significant data cleaning is done for FER-2013 dataset.
    
\begin{figure}[!h]
        \centering
        \includegraphics[scale = 0.35]{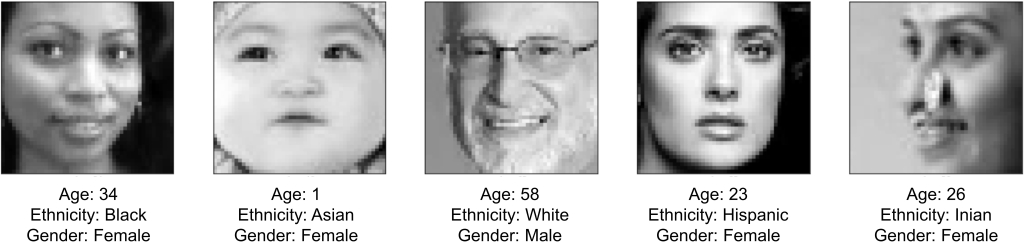}
        \caption{Samples from the AGE, GENDER AND ETHNICITY dataset \cite{ageData}}
        \label{fig:eth_age_sample}
    \end{figure}

     \vspace{-0.3cm}
    The second goal is the detection of age and ethnicity by using AGE dataset (Fig. \ref{fig:eth_age_sample}). There are in total 5 columns: age, ethnicity, gender, image name and pixels. Since gender labels are not used in the developed system due to the lack of contribution of it, they are erased. The pixels represent the image, yet they need to be preprocessed to be visualized and trained. To do that, firstly they need to be transformed from string into float datatype and gathered up in Numpy arrays. Then, they are reshaped into 2 dimensions. 
    
    \vspace{-1cm}
    \subsubsection{Normalization}
    Normalization, also known as feature scaling, is a method of placing the value of a feature within a certain range. Scaling is commonly done in the range of 0 to 1. For example; feature A can range from 0 to 80, while feature b can range from 15000 to 25000. When all these values are normalized by compressing them between a certain range, especially distance-based machine learning algorithms are known to give better results in general, but may not always give better results. For this reason, it should definitely be tried and used if it is observed that it gives good results. 
    In this case, according to the observation obtained, normalization of the pixels has increased the efficiency of the model.
    
    \subsection{Proposed CNN Architecture}
    
\begin{figure}[!h]
        \centering
        \includegraphics[scale = 0.50]{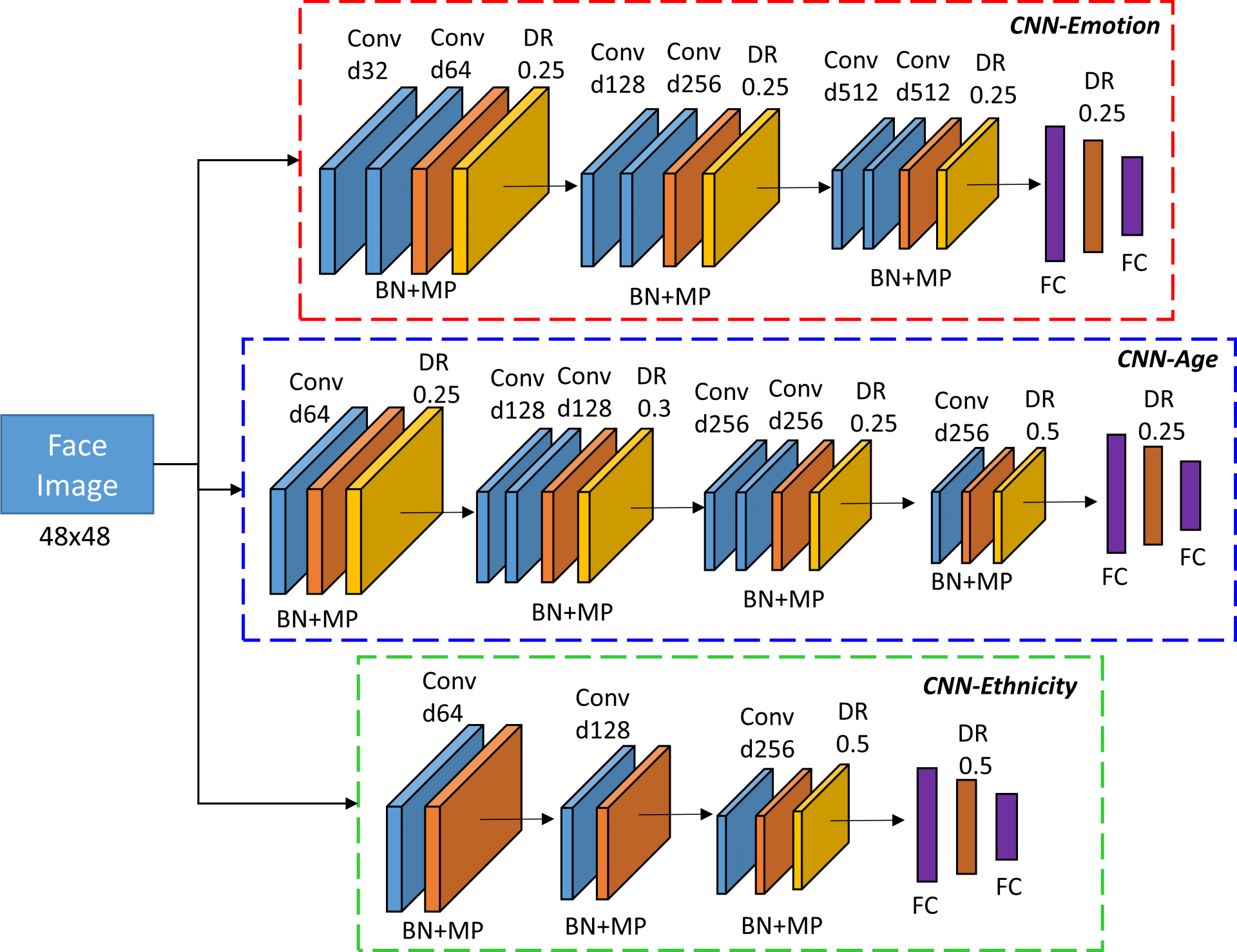}
        \caption{CNN Architectures used in the Proposed Method}
        \label{fig:CNN}
    \end{figure}

    CNN is a very popular deep learning method used in many applications. Generally, it can be developed by using different combinations of five different layer types: i) convolutional (Conv), ii) max-pooling (MP), iii) batch normalization (BN), iv) dropout (DR), and v) fully connected (FC)layers. In Conv layer, various filters are navigated by applying convolution operation on the image. The purpose of this process is to try to understand the important features of the image in the background. The MP layer reduces the size of the image in a certain way, reducing the amount of processing. It also helps the model to get rid of unnecessary information. On the other hand, large number of features can cause overfitting. In order to tackle these problem, DR layer can be used such CNN architectures. The FC layer, which is generally the last layer, is used to perform the classification process by optimizing its weights as a result of transforming the image into a vector after shrinking and giving it as an input to the FC layer. In the proposed method, we have used all these types for the designed CNN architecture (Fig. \ref{fig:CNN}). 
  \vspace{-0.2cm}
       
    For the prediction of emotion, age and ethnicity, different CNN models are used in this paper. To find the best layer ordering and the best hyperparameters of these layers, we have tried many combinations. We have found that using two cascaded Conv layers with increasing filter (e.g., d32 and d64) sizes gives better results for the prediction of emotion. Also, DR enables to prevent overfitting and provides better classification rates. On the other hand, determination of ethnicity type looks a simpler problem than emotion, and shallower architecture with lower number of layers can give satisfying results here. Due to the age range changes from 0 to 110 values, there are many labels to be predicted. Therefore, deeper layers used to predict the age results efficiently. Additionally, mean square error loss is utilized for the optimization of CNN-Age network whereas categorical cross entropy is used for CNN-Emotion and CNN-Ethnicity newtorks.

    The hyperparameters to be known in CNN architecture are kernel size, padding and activation. Kernel size, which basically is convolutional filter is decided to be in 3 x 3 matrix size for all Conv layers. ReLu is used as an activation function. Padding, on the other hand, is a term relevant to convolutional neural networks as it refers to the amount of pixels added to an image when it is being processed by the kernel of a CNN. 
    
\section{Web Application Phase}
\subsection{.Net Technologies}
.NET is a free, cross-platform, open source developer platform for building many
different types of applications which developed by Microsoft. First of all, let’s talk about .Net-based application models. With the .Net Framework, which was the first
of these and used more because it is old, projects that could only run on Windows
could be produced. For this reason, .Net Framework, which is not preferred by Mac
OS and Linux users, has emerged with a brand new version. With this version called
.Net Core, projects that can run on all these platforms can now be produced. On the other hand ASP .NET Core is an open source framework that allows to build cloud-based applications such as web servers, IoT and mobile API’s/backends.
\subsection{Model View Controller}
MVC is a software architecture includes three parts called Model, View and Controller.

\textbf{Controller:} Controller part of the MVC is responsible from all the business layer,
processes between view and model, database queries and view triggers.

\textbf{Model:} Model part gives the structure of the datas to be used in the project. This
models could be used and transferred over the views and the controllers.

\textbf{View:} View includes the frontend of the project. It accepts the data returns from the controller.
\subsection{Flask}
Flask is a Framework that is developed for serving Python Programs as a server. Basically, Flask gives the opportunity to build an API and create an endpoint for a
Python script. In this project, Flask creates APIs for each of the three models used for classification of the age, ethnicity and emotion. Those flask servers returns the output of the models. Flask provides to the all models work independently from each other.
\subsection{Spotify API’s}
Spotify provides it’s APIs to it’s payed customers under the Developer Accounts. It provides some information like tracks, playlist, albums ex. To access APIs, an
Authentication required. To create an access token, account information should be used.
\section{Application}
In this section, application steps are explained one by one in the following paragraphes:  

 \textbf{Creating an MVC Project:} Web Site of the project is created using MVC architecture with ASP .Net Core application. To create this type of project, Visual Studio 2019 Integrated Development Environment is used. There are three Controllers and three Views which creates 3 different pages in the MVC project.
 
 \textbf{Capture The Photo From The Computer Camera Device:} First step of the development is capturing the photo from the camera of the computer. A controller named CameraController is created for this purpose. The captured photo is saved to the given path.
 
 \textbf{Create API From Models With Flask:} To use the models on the captured photo, there is a need to create APIs for the models. Best way to generate API for Python scripts is using Flask. It also provides endpoints to access the output. There is not any model education in these APIs. A photo is sent to the models and an output is taken from the endpoint.
    \begin{figure}[!h]
        \centering
        \includegraphics[scale = 0.50]{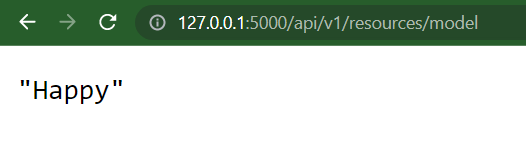}
        \caption{Flask Endpoint Result}
        \label{fig:flask}
    \end{figure}
    \vspace{-0.2cm}

 \textbf{Preprocessing of The Captured Photo:} To send the captured photo to the models, the snapshot needs to be cropped as just face. An OpenCV library called HaarCascade crops the photo the way needed. The python method saves the cropped face when the model API is used.
    \begin{figure}[!h]
        \centering
        \includegraphics[scale = 0.70]{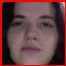}
        \caption{Cropped Face Created Using HaarCascade}
        \label{fig:cropped}
    \vspace{-0.2cm}
    \end{figure}
    
 \textbf{Connect Flask APIs From MVC Architecture:} The operation of connecting to Flask API is done by the MVC project which is developed in .Net Core. As described above, all business operations are done in the
Controller part of MVC Architecture. Since sending web request is also a business layer operation, it is done in the Model Controller.

 \textbf{Creating Playlist Data:} Due to three classifier outputs, there is 80 different possible playlist to suggest. Controlling all the possible values by using conditional statements is not an effective and possible way to find the proper playlist to suggest. To solve this problem, a JSON is created to access the right playlist ID. JSON file is implemented into the project and accessed via the MVC project. All the 80 playlists are chosen to provide the proper musics for the user. 
 
 \textbf{Authentication and Send Request to Spotify APIs:} To connect to a Spotify API, an authentication must be fulfilled. For that, the controller should generate a Spotify Client by sending an access token. After the connection is fulfilled, it is possible to take all the playlist or tracks needed. In this project, it is preferred to send ’Get Playlist’ request for all the possible combinations from the three models’ outputs. 

 \textbf{Creation of Frontend Using Views:} In MVC architecture, for all of the pages in the web site, there must be a view file. View file basically contains HTML document. It is possible to add CSS and Java Script files to be used by the view.
\subsection{Flow}
In this section, the flow of the web site is explained. There are three pages in the website. These are the Home Page, the Capture Image Page and the Suggest Song Page.

 \textbf{Home Page:} Home Page is the default page which appears when the application started. It gives the basic information about the web site and the authors. 
 
 \textbf{Capture Photo Page:} This page provides the snapshot feature to the user. The captured image has been sent to the machine learning models which have turned into services. Snapshot process happens after clicking to ’Take Snapshot’ button. The captured photo is displayed above the camera.
 
 \textbf{’Suggest Music’ Page:} The final page is the page which the song suggestions are displayed. Musics of the matching playlist is listed according to the results of the three classifiers. The playlist id is chosen from the JSON which includes playlist IDs for all the possible scenarios. To list the songs, Bootstrap Table is used. All the songs, artists and albums in the table are clickable links which directs the user to the Spotify Web Page for the clicked element. This feature provided by parsing a link for the items. Due to Spotify API Get Requests, the program is able to access the ID of the tracks, albums and the artists.

 \vspace{-0.5cm}
 
     \begin{figure}[!h]
        \centering
        \includegraphics[width=\linewidth, frame, scale = 0.8]{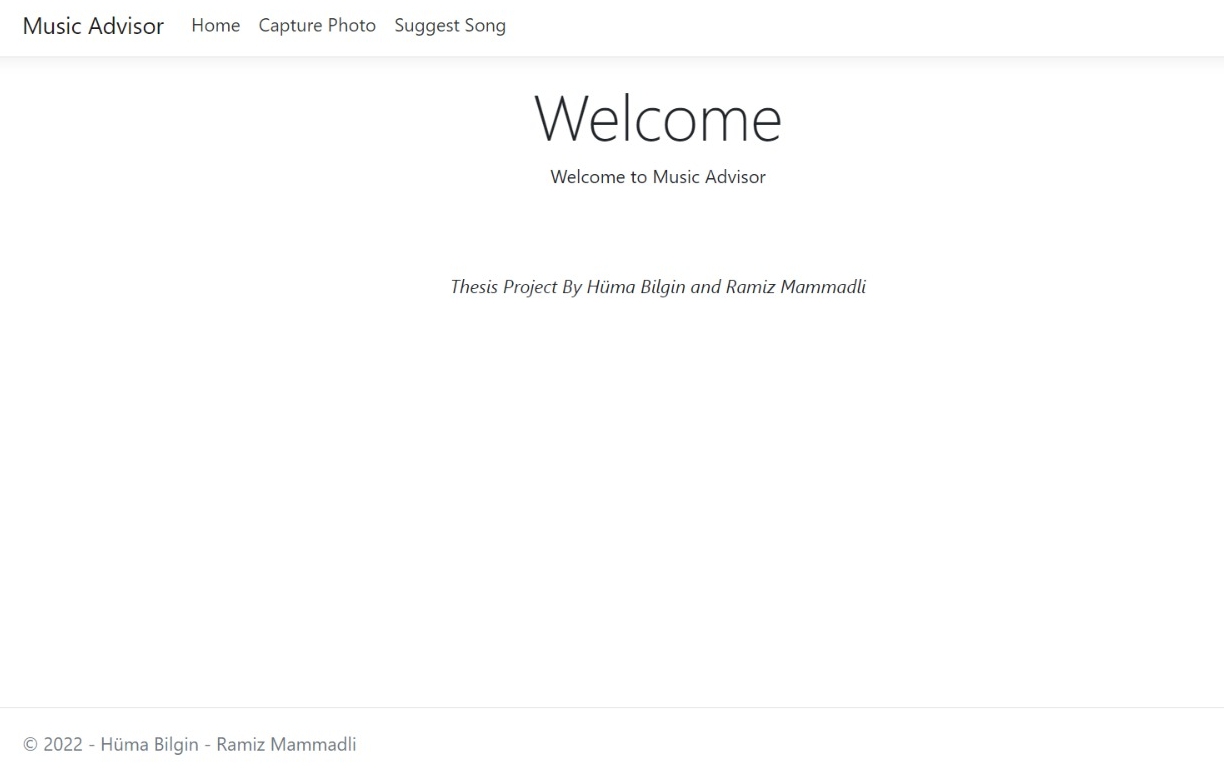}
        \caption{Home Page}
        \label{fig:homepage}
    \end{figure}
    
 \vspace{-1.2cm}
 
   \begin{figure}[!h]
        \centering
        \includegraphics[width=\linewidth, frame, scale = 0.45]{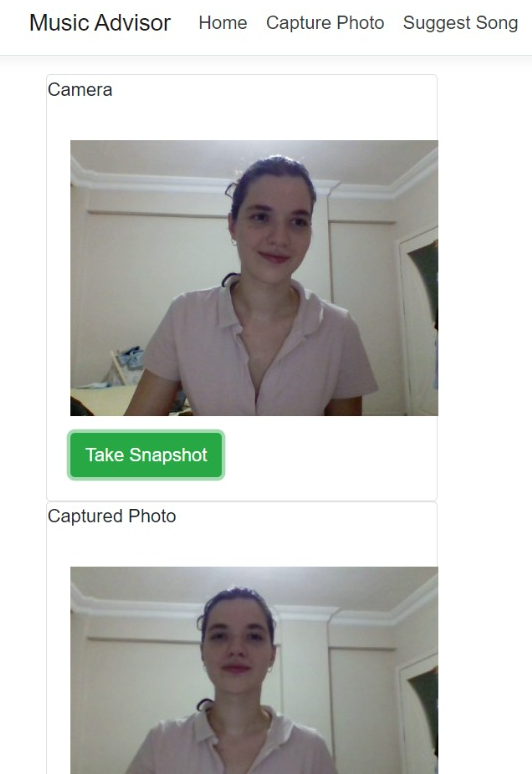}
        \caption{Capture Photo Page}
        \label{fig:capturephoto}
    \end{figure}

    \begin{figure}[!h]
        \vspace{-0.4cm}
        \centering
        \includegraphics[scale = 0.27]{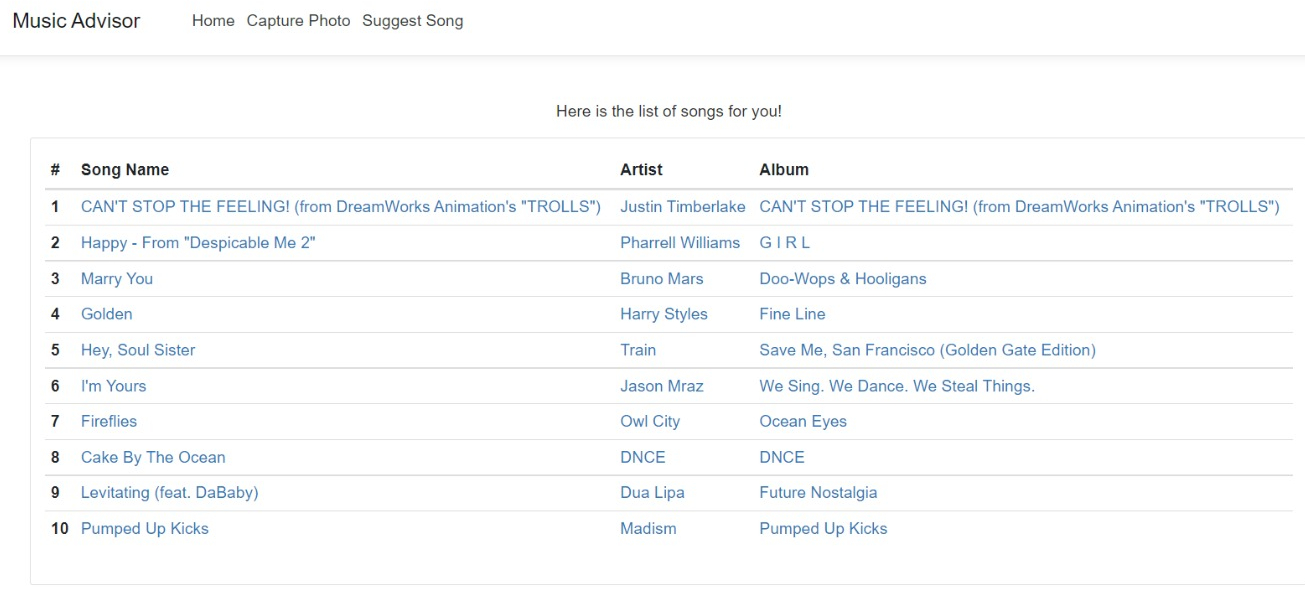}
        \caption{Suggest Music Page}
        \label{fig:suggestmusic}
    \end{figure}

\section{Experimental Results}
    \subsection{Results for FER-2013 dataset}
    In order to determine the best combination, we have tried different number of convolutional layers and pooling layers for FER-2013 dataset.  The results are compared in terms of F1-score and classification accuracy and they are shared in Table 1. According to the table, the best combination pair is obtained when the number of convolution and pooling layers are 6 and 3, respectively. Selecting this pair provides F1-score of 0.7497 and the accuracy of 0.8125. 
    Note that we have developed the CNN models as simpler as possible because of the computational complexity. Simpler models allow us to fast prediction of emotion, age and ethnicity categories.
   
    \subsection{Results for Age, Gender and Ethnicity dataset}
    
    Since this dataset has a similar structure with FER-2013, the architecture used in the previous model has been decided to be used here as well. The differences can be seen in Fig. 4. Here, CNN-Age and CNN-Ethnicity are two models with different structures. One of them is scattered into the range between 0 and 116, in numbers, while the other is just multiclass data. Therefore, it is preferred to train 2 separate models by taking the subdataset out of the main one. 
	Similar to the previous subsection, F1-scores and the clasification results of several combinations including number of convolutional and pooling layers are obtained for CNN-Age and CNN-Ethicity models, which can be found in Table 2 and Table 3, respectively. Due to using mean square error (MSE) as loss function, the results of CNN-Age are given in terms of mean square error and mean absolute error metrics. In addition, metrics used in CNN-Ethnicity model is the same with CNN-Emotion. From these experiments, we see that deeper model is needed for the prediction of age whereas shallower model is enough for the classification of ethnicity.

    \begin{table}[!h]
\begin{tabular}{|lllll|}
\hline
\multicolumn{1}{|l|}{Num. of trial} & \multicolumn{1}{l|}{Num. of Conv. Layers} & \multicolumn{1}{l|}{Num. of Pooling layers} & \multicolumn{1}{l|}{F1-score} & Accuracy \\ \hline
\multicolumn{1}{|l|}{1}          & \multicolumn{1}{l|}{5}                         & \multicolumn{1}{l|}{2}                   & \multicolumn{1}{l|}{0.6126}   & 0.6725   \\ \hline
\multicolumn{1}{|l|}{2}          & \multicolumn{1}{l|}{5}                         & \multicolumn{1}{l|}{3}                   & \multicolumn{1}{l|}{0.6648}   & 0.7687   \\ \hline
\multicolumn{1}{|l|}{3}          & \multicolumn{1}{l|}{6}                         & \multicolumn{1}{l|}{3}                   & \multicolumn{1}{l|}{0.7497}   & 0.8125   \\ \hline
\multicolumn{1}{|l|}{4}          & \multicolumn{1}{l|}{6}                         & \multicolumn{1}{l|}{4}                   & \multicolumn{1}{l|}{0.7005}   & 0.7112   \\ \hline
\multicolumn{1}{|l|}{5}          & \multicolumn{1}{l|}{7}                         & \multicolumn{1}{l|}{5}                   & \multicolumn{1}{l|}{0.6031}   & 0.6629   \\ \hline
\end{tabular}
 \vspace{0.2cm}
\caption{\label{tab:Table I.}Results of CNN-Emotion model in various layer combinations}

\begin{tabular}{|llllll|}
\hline
\multicolumn{1}{|l|}{Num. of trial} & \multicolumn{1}{l|}{Num. of Conv. Layers} & \multicolumn{1}{l|}{Num. of Pooling layers} & \multicolumn{1}{l|}{F1-score} & \multicolumn{1}{l|}{MSE}     & MAE     \\ \hline
\multicolumn{1}{|l|}{1}          & \multicolumn{1}{l|}{5}                         & \multicolumn{1}{l|}{3}                   & \multicolumn{1}{l|}{0.5528}   & \multicolumn{1}{l|}{88.9334} & 10.2689 \\ \hline
\multicolumn{1}{|l|}{2}          & \multicolumn{1}{l|}{6}                         & \multicolumn{1}{l|}{3}                   & \multicolumn{1}{l|}{0.6022}   & \multicolumn{1}{l|}{79.0293} & 7.9102  \\ \hline
\multicolumn{1}{|l|}{3}          & \multicolumn{1}{l|}{6}                         & \multicolumn{1}{l|}{4}                   & \multicolumn{1}{l|}{0.6528}   & \multicolumn{1}{l|}{62.4204} & 5.5289  \\ \hline
\end{tabular}
 \vspace{0.2cm}
\caption{\label{tab:Table II.}Results of CNN-Age model in various layer combinations}

\begin{tabular}{|lllll|}
\hline
\multicolumn{1}{|l|}{Num. of trial} & \multicolumn{1}{l|}{Num. of Conv.Layers} & \multicolumn{1}{l|}{Number of Pooling layers} & \multicolumn{1}{l|}{F1-score} & Accuracy \\ \hline
\multicolumn{1}{|l|}{1}          & \multicolumn{1}{l|}{2}                         & \multicolumn{1}{l|}{2}                   & \multicolumn{1}{l|}{0.5528}   & 69.62    \\ \hline
\multicolumn{1}{|l|}{2}          & \multicolumn{1}{l|}{3}                         & \multicolumn{1}{l|}{3}                   & \multicolumn{1}{l|}{0.6022}   & 73.74    \\ \hline
\multicolumn{1}{|l|}{3}          & \multicolumn{1}{l|}{3}                         & \multicolumn{1}{l|}{3}                   & \multicolumn{1}{l|}{0.6928}   & 80.74    \\ \hline
\end{tabular}
 \vspace{0.2cm}
\caption{\label{tab:Table III.}Results of CNN-Ethnicity model in various layer combinations}
  \vspace{-0.6cm}
\end{table}

\subsection{Some Application Results}

\begin{figure}[!h]
 \vspace{-0.2cm}
        \centering
        \includegraphics[scale = 0.45]{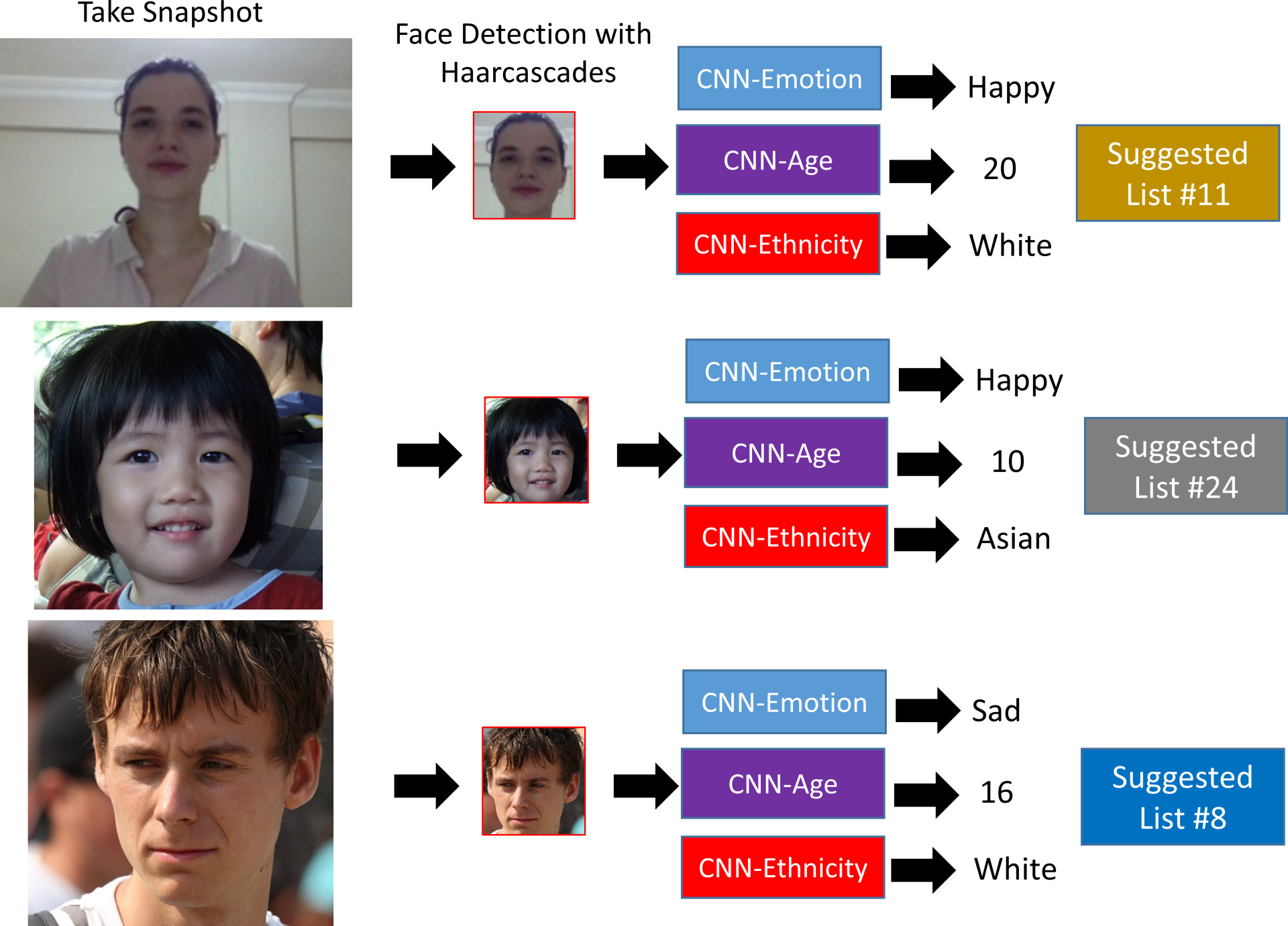}
        \caption{Results of the Proposed System for Three Different People}
        \label{fig:flow}
    \end{figure}

In this subsection, three different image inputs are feeded to the designed CNN-architectures and obtained results of them are shown in Fig. \ref{fig:flow}. 

Here, the first image is the image of the second author captured with webcam, and the second and the third one are used from FFHQ dataset \cite{ageData}. It is seen that emotion, age and ethnicity results are predicted well. For example, the real age of the first person is 22 whereas it is predicted as 20. Due to each photo has emotional, age or ethicity differences from each other, we suggest different lists from Spotify API.

\section{Performance Analysis}
    The model first planned to be trained by using transfer Learning methods. Transfer learning is the methodology where the initial parameters and architectures of the models that have been trained and proven successful before with very large data are taken and the layers are frozen. Then, FC layers are added to the end of the model and fine tuning is performed with the data set planned to be used in the project. In this way, both the initial parameters get rid of randomness and start with good values, and since only the last FC layers will be trained, time and computational power are saved. However, transfer learning model was trained on the images which have both different size and color channel than the dataset used.
    
    Nonetheless, transfer learning method is not preferred. Because, both in FER-2013 and AGE datasets, the images are in grayscale, which means they have only one color channel. On the other hand, data used to train the transfer learning methods has 3 color channels. For these reasons, FERNet-like architecture is used to train the model, which is specifically designed for FER-2013 dataset, based on CNN architectures. Since the second dataset has the images similar to FER-2013, same architecture has performed a fulfilling outputs. 
    
    As a result, the accuracy of the model on test data has reached 81.25\% in FER-2013. Getting this high accuracy has satisfied the goal accuracy, set before starting the project for this dataset. 
    
    The accuracy of the age model is measured by using MAE, which is roughly 5.53. In other words, to calculate the accuracy, the ratio of the mean age of dataset and MAE can be taken. The mean age is 33, makes the ratio about 0.167,  which is about 17\%. To use the age classifier in the music recommendation system,  output is quite optimistic, since the age is just going to help the system to categorize users to propose relative playlists.
    
    After the completion of the Ethnicity model, the accuracy has been reached 81.46\%. Considering there are 5 different ethnicity groups, and the grayscale characteristic of the dataset, it can be noted that in practice this helps the system to classify people into ethnic origins.
    
    Moreover, in the models where highest accuracy or MSE/MAE has been achieved, F1 score is in the greatest level as well. Therefore it can be justified that models implemented to the system as classifiers are the highest performers among their peers.  
    
\section{Conclusion}
    In this project, a system that makes an automatic music playlist recommendation system based on emotion, ethnicity and age on images taken over the web application has been developed. It has first been outlined that the transfer learning methods would be useful. Due to the reasons mentioned in Performance Analysis, more suitable architecture based on CNN is used. The trained models are saved and integrated to the web application-end of the project. Here, the models are used as classifiers by using the snapshot of the user taken via interface. According to the mutual output of 3 separate models, the playlist will be proposed to the user to build a pleasing user experience. To create a web site .Net technologies have been used. MVC Architecture was the main software of the project. To use generated models, they are turned into an API with Flask. By sending request to Spotify APIs, the proper song list has shown to the user.
    
    After the development of the project, it is observed that all the trained models are working successfully in collaboration with the Web Application. All these features lead to the fulfilling result aimed in the plan of the system.
  
%
%
%
%

\end{document}